\documentclass[10pt,twocolumn,letterpaper]{article}

\usepackage{cvpr}              

\usepackage{graphicx}
\usepackage{amsmath}
\usepackage{amssymb}
\usepackage{booktabs}
\usepackage{pifont}
\newcommand{\cmark}{\ding{51}}%
\newcommand{\xmark}{\ding{55}}%
\usepackage{multirow}

\usepackage[pagebackref,breaklinks,colorlinks]{hyperref}

\usepackage[capitalize]{cleveref}
\crefname{section}{Sec.}{Secs.}
\Crefname{section}{Section}{Sections}
\Crefname{table}{Table}{Tables}
\crefname{table}{Tab.}{Tabs.}

\def\cvprPaperID{16} 
\def\confName{CVPR}
\def\confYear{2023}

\begin{document}

\title{Isolated Sign Language Recognition based on Tree Structure Skeleton Images}

\author{David Laines*, Miguel Gonzalez-Mendoza, Gilberto Ochoa-Ruiz*\\
Tecnologico de Monterrey, School of Sciences and Engineering\\
Av. Eugenio Garza Sada 2501 Sur, Tecnológico, 64849 Monterrey, N.L.\\
{\tt\small *Corresponding: davidlainesv@outlook.com, gilberto.ochoa@tec.mx}
\and
Gissella Bejarano\\
Universidad Peruana Cayetano Heredia\\
Av. Honorio Delgado 430, Urb Ingeniería, Lima, Perú
{}
}
\maketitle

\begin{abstract}
   \vspace{-3mm}
   Sign Language Recognition (SLR) systems aim to be embedded in video stream platforms to recognize the sign performed in front of a camera. SLR research has taken advantage of recent advances in pose estimation models to use skeleton sequences estimated from videos instead of RGB information to predict signs. This approach can make HAR-related tasks less complex and more robust to diverse backgrounds, lightning conditions, and physical appearances. In this work, we explore the use of a spatio-temporal skeleton representation such as Tree Structure Skeleton Image (TSSI) as an alternative input to improve the accuracy of skeleton-based models for SLR. TSSI converts a skeleton sequence into an RGB image where the columns represent the joints of the skeleton in a depth-first tree traversal order, the rows represent the temporal evolution of the joints, and the three channels represent the (x, y, z) coordinates of the joints. We trained a DenseNet-121 using this type of input and compared it with other skeleton-based deep learning methods using a large-scale American Sign Language (ASL) dataset, WLASL. Our model (SL-TSSI-DenseNet) overcomes the state-of-the-art of other skeleton-based models. Moreover, when including data augmentation our proposal achieves better results than both skeleton-based and RGB-based models. We evaluated the effectiveness of our model on the Ankara University Turkish Sign Language (TSL) dataset, AUTSL, and a Mexican Sign Language (LSM) dataset. On the AUTSL dataset, the model achieves similar results to the state-of-the-art of other skeleton-based models. On the LSM dataset, the model achieves higher results than the baseline. As far as we know, our work is the first to try TSSI for sign language recognition and our results suggest it presents a real alternative for isolated sign language representation. Code has been made available at: \textcolor{blue}{https://github.com/davidlainesv/SL-TSSI-DenseNet}.
\end{abstract}

\section{Introduction}
\label{sec:introduction}

\begin{figure*}[ht]
 \centering
 \includegraphics[width=\linewidth]{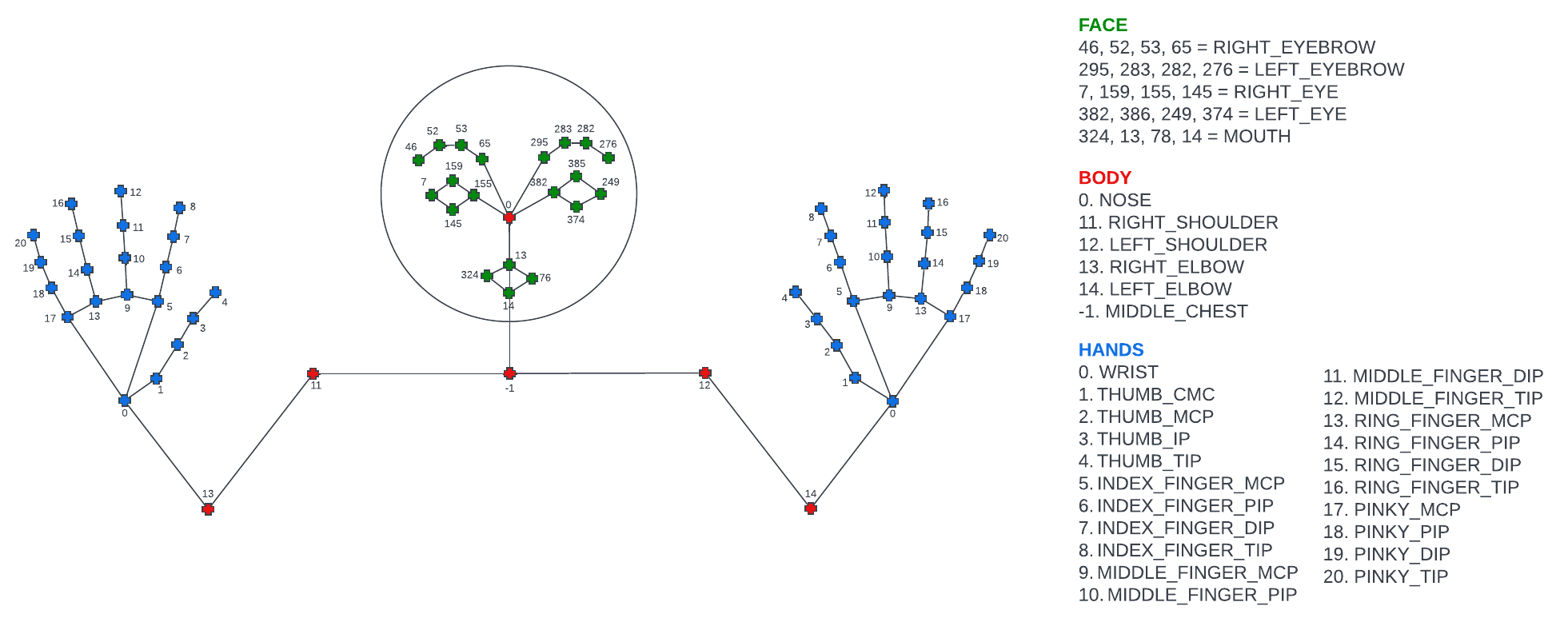}
 \caption{Selection of keypoints and connections taken from the MediaPipe holistic model to generate a base skeleton graph. The keypoint indicated with -1 represents the inner chest calculated at the middle of the shoulders.}
 \label{fig:skeleton_graph_representation}
\end{figure*}

In recent years, we have witnessed enormous progress in the domain of Human Action Recognition (HAR), including sports analysis, video surveillance, and sign language recognition (SLR), among many others. SLR is an important task not only from a technical point of view but also from a social perspective. Currently, the deaf community does not have equal access to all areas of society due mainly to the lack of sign language users \cite{haualand2009sign}. Isolated Sign Language Recognition (ISLR) is an instance of SLR that aims to map an isolated sign language video into a word of a written language or a \textit{gloss}. Robust ISLR models can be embedded in search engines to support sign language and self-paced teaching frameworks to make sign language acquisition easily accessible to hearing people when sign language teachers are not available.


In the last decade, deep learning (DL) methods such as 3D convolutional neural networks (3DCNNs) have been proposed to tackle ISLR \cite{rastgoo2020video, hosain2021hand}. 3DCNNs directly map the RGB data of a video into a label. However, 3DCNNs usually require a large number of parameters to obtain dimensional representations, which increases the computational complexity of the model \cite{pham2019learning}. This is not suitable in scenarios where the model is required to run directly on mobile devices to avoid sending private information to the internet.

Skeleton-based features can be used to produce less complex HAR models while being robust to changes in the background, lightning conditions, and physical appearance \cite{yue2022action}. To process skeleton-based features, graph neural networks (GNN), convolutional neural networks (CNN), and Transformers have been proposed. CNN-based methods usually encode a skeleton sequence into an image which allows the use of 2D convolutional neural networks (2DCNN) to process the image. GNNs represent the spatio-temporal characteristics of a skeleton sequence as a graph data structure and directly operate on it whereas Transformers take a skeleton sequence as a single vector. Previously GNNs and Transformers have been applied to ISLR \cite{de2019spatial, vazquez2021isolated, bohavcek2022sign}, but there is still little research applying CNN-based methods.

In this work, we propose the use of the Tree Structure Skeleton Image (TSSI) method \cite{yang2018action} to represent the spatio-temporal characteristics of isolated sign language skeleton sequences. 
In contrast to conventional HAR, the body pose joints are not sufficient to classify sign language signs. As noted by \cite{von2008significance}, facial expression has an important significance in SLR, as much as the hands. We extend the work of \cite{yang2018action} to take into account not only the body pose joints but also the facial and fine-grained hand joints in the construction of the base skeleton graph that is required for generating TSSI images. Moreover, we process the generated TSSI images with a less complex deep learning network than the originally proposed, the Dense Convolutional Network (DenseNet) \cite{huang2017densely}, a very well-known 2DCNN network for image classification. Finally, we evaluate our approach with a large-scale and popular video dataset for SLR, the WLASL dataset. Additionally, we evaluate the effectiveness of our approach on the AUTSL dataset and a publicly available Mexican Sign Language (LSM) dataset \cite{mejia2022automatic}. To estimate the skeleton data from the sign language video datasets, we decided to use MediaPipe \cite{lugaresi2019mediapipe}, a pose estimation model that estimates human pose along with hands and facial keypoints from video frames. Hereinafter we use joints and keypoints interchangeably.

The main contributions of this paper are:
\begin{itemize}
    \item Applying the TSSI method, for the first time, to 3 ISLR datasets, WLASL, AUTSL, and an LSM dataset.
    \item Proving that TSSI brings competitive results when used with a DenseNet while keeping a low number of parameters in comparison to other skeleton-based models and RGB-based models.
\end{itemize}

The rest of the paper is organized as follows. In Section \ref{sec:related_work}, we compare previous works on ISLR and describe their main characteristics. In Section \ref{sec:methodology} we introduce the proposed skeleton-based sign language representation. In Section \ref{sec:experiments_and_analysis} we describe the selected dataset used in our experiments, present the experimental setup and provide an analysis of the results. Finally, in Section \ref{sec:conclusions} we present the conclusions of our work.

\section{Related Work}
\label{sec:related_work}
In this section, we review previous works for HAR that represent skeleton sequences as images and process the images with convolutional neural networks (CNNs). These methods are categorized in the literature as CNN-based methods.

In \cite{du2015skeleton}, one of the first CNN-based works, the joints of a human skeleton are divided into 5 groups, i.e., left arm, right arm, left leg, right leg, and trunk. The joints are listed together in a vector of joints $v$ of length $N$ that holds a new order of joints sorted by group. Consequently, a skeleton sequence of $T$ frames is represented as an RGB image $I$ such that $I=[p_{1,1}, p_{i,j}, ..., p_{N,T}]$ where $N$ indicates the total number of joints, $T$ indicates the total number of frames, and $p_{i,j}$ indicates the pixel describing the $(x, y, z)$ coordinates in the $(r, g, b)$ channels, respectively, of the joint at $v_i$ and frame at index $j$.

\begin{figure*}[ht]
 \includegraphics[width=\linewidth]
 {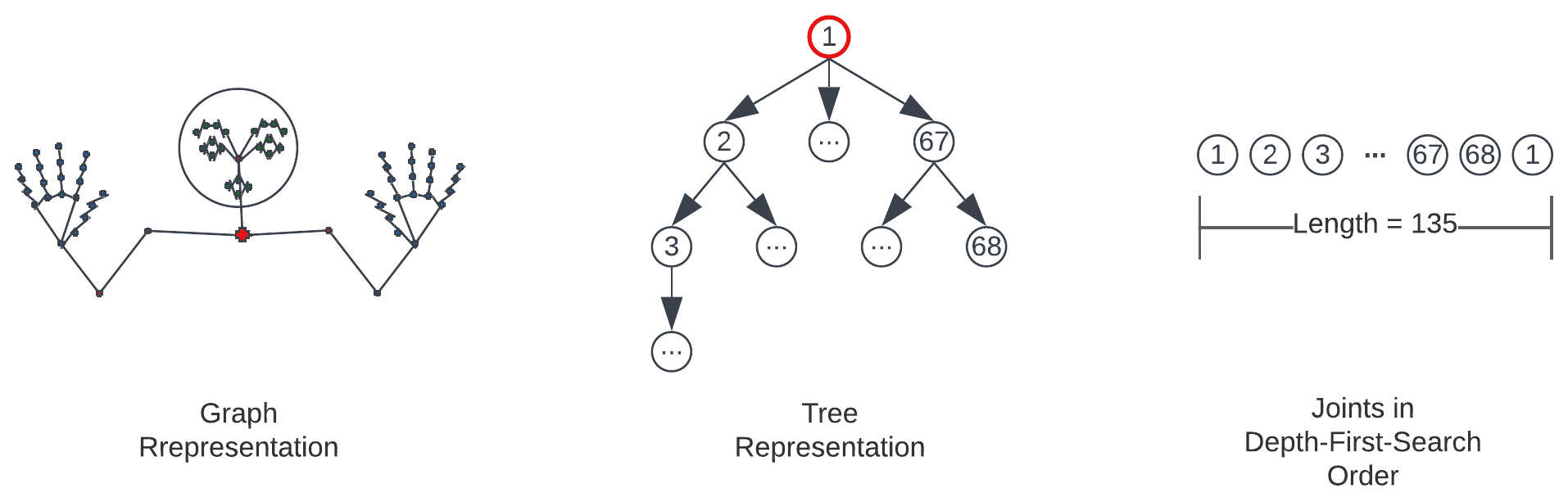}
 \centering
 \caption{A base skeleton graph (left) is used to build a tree structure (middle) starting from the root joint (in red). DFS is applied to the tree to get an ordered list of joints (right).}
 \label{fig:joints_order}
\end{figure*}

As convolutional operations aggregate neighboring pixels across the layers of a CNN, \cite{li2018co} proposed a model architecture that maps the skeleton joints in the channel dimension of the layers of the network to preserve the relation between all joints for longer sequences. In \cite{banerjee2020fuzzy}, rather than using the raw coordinates of the joints to represent the pixels of an image for further classification, the authors use instead the values of the angles and the distances between subsets of joints across the frames.

Other works opted for fusing the raw coordinates with additional measures obtained from the skeleton joints. For example, in \cite{li2018co}, the authors fuse the output of processing an image generated using the raw coordinates with another image generated using the temporal difference of the joints between consecutive frames. In \cite{wang2021skeleton} they fuse the raw coordinates with a heatmap representation of the joints.

A better selection of the order of the joints is important to better exploit the mathematical properties of the convolutional operations made by the CNNs. \cite{yang2018action} proposed the Tree Structure Skeleton Image method that traverses a naive skeleton graph with depth-first-search (DFS) to find an order of joints that represent the connections among them. The order of the joints in the resulting path is represented as the columns of the image and the temporal evolution of the joints is represented as rows.

In \cite{nunez2018convolutional} they noted that augmentation functions can improve the performance of a model. In this work, they use 5 augmentation functions: scale, shift, noise, subsample, and time interpolation. Their experimental results on datasets of different sizes showed that the augmentation functions have a greater impact on small datasets due to their capability to extend the diversity of the training set during the learning process of a model.

Very recently, \cite{duan2022revisiting} used 3D heatmap volumes to represent skeleton sequences and a lightweight version of a 3D-CNN to process the input. They outperformed GNN-based methods in performance, robustness, and efficiency across different HAR datasets. However, in contrast to GNNs, the CNN-based approach has not been fully explored for Sign Language Recognition.

\section{Methodology}
\label{sec:methodology}
Herein, we describe in detail the components of the proposed representation.

\subsection{Order of joints}

\label{sec:order_of_joints}
A base human skeleton was represented by a graph $G = (E, V)$ where $V$ is the set of nodes w.r.t. skeletal joints, and $E$ is the set of edges that connect the nodes where $e = (v_i, v_j), e \in E$ and $v_i, v_j \in V$. The nodes and the edges of the graph were designed based on the physical structure used by the MediaPipe holistic model. This model locates 543 keypoints representing the joints of the human body and defines their connections. We selected a subset of 68 keypoints based on the work of \cite{mejia2022automatic}. As shown in Figure \ref{fig:skeleton_graph_representation} the joints include 6 body keypoints, 20 face keypoints, 21 left-hand keypoints, and 21 right-hand keypoints. Due to its importance \cite{von2008significance}, the face keypoints include four points for each eyebrow, four points around each eye, and four points around the mouth.

As shown in Figure \ref{fig:joints_order}, we took the base skeleton graph to build a tree structure taking the joint at the middle of the shoulders as the root node. Subsequently, we performed Depth-First-Search (DFS) over the tree to get a path of joints in the order they were visited. We obtained an ordered list of joints with a length of 135. We used this ordered list of joints to generate a Tree Structure Skeleton Image (TSSI) in further steps.

\begin{figure*}[ht]
 \includegraphics[width=\linewidth]{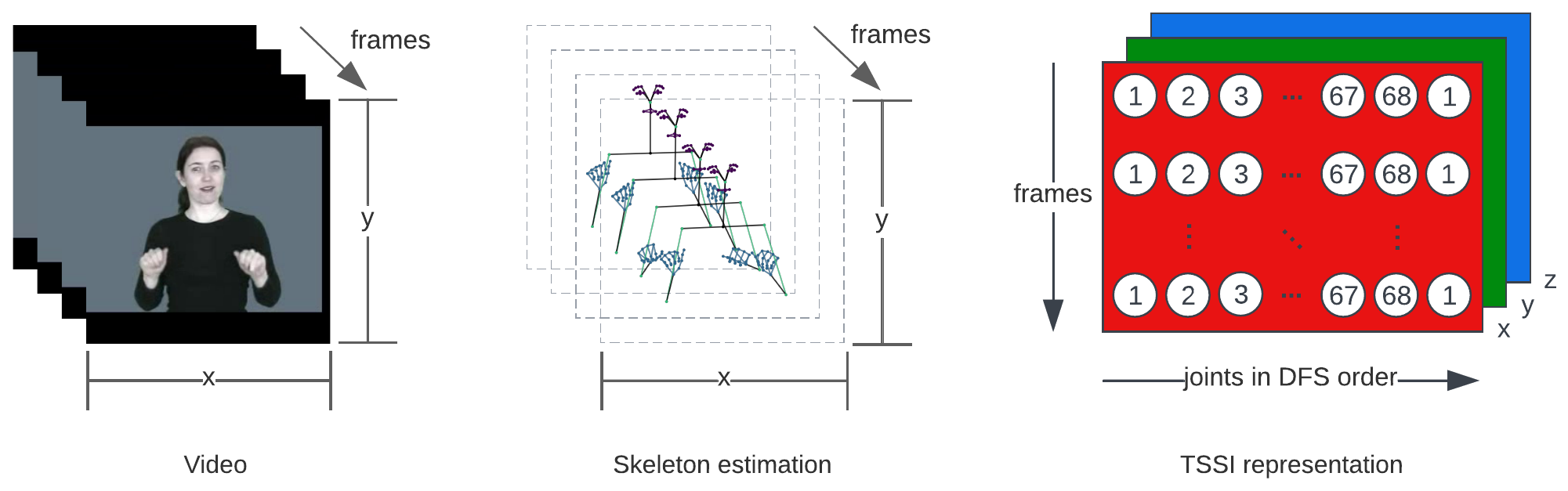}
 \centering
 \caption{Process to generate a TSSI representation from a video. The skeleton data is extracted with MediaPipe at every frame of the video. Then, an image is generated such that the joints are placed in DFS order in the column dimension, the frames are stacked in the row dimension, and the coordinates (x, y, z) of the joints in the channel dimension.}
 \label{tssi_representation}
\end{figure*}

\subsection{TSSI representation}
Given a video, we extracted the skeleton data at each frame with MediaPipe and used the path of joints obtained by DFS to generate a Tree Structure Skeleton Image (TSSI) as shown in Figure \ref{tssi_representation}. The skeleton data was normalized to [0.0, 1.0] by the image width and height, respectively. We discarded the frames where MediaPipe could not estimate the body pose. When MediaPipe failed to estimate the left and right hand coordinates, we replaced those coordinates with the coordinates of the wrist. When MediaPipe failed to estimate the face coordinates, we replaced those coordinates with the coordinates of the nose.

In a TSSI, the rows contain the skeleton data of every frame in the video, the columns contain the skeleton data of every joint of the skeleton in the order they were visited by DFS and the channels $(r, g, b)$ contain the $(x, y, z)$ coordinates of the joints, respectively. Specifically, given an ordered list of joints $v$ (described in Section \ref{sec:order_of_joints}) of length $N$ and a video with a total number of frames $T$, a TSSI representation $I$ is generated such that $I=[p_{1,1}, p_{i,j}, ..., p_{N,T}]$ where $p_{i,j}$ indicates the pixel describing the $(x, y, z)$ coordinates in the $(r, g, b)$ channels, respectively, of the joint at $v_i$ and frame at index $j$.

As the documentation of the MediaPipe holistic model advises that, currently, the estimation of the z-axis is not reliable, we opted for setting the blue channel to 0 in all the pixels to form a full RGB image. Due to that videos can have varying lengths, we resized the generated images to a uniform size of 135xH (width x height) where H is a fixed length based on the mean length of the sequences in the training set. If the height of the resulting TSSI was greater than H pixels, we resized it using bilinear interpolation. On the other hand, if it was smaller, we padded the image with zeros.


\section{Experiments and Analysis}
\label{sec:experiments_and_analysis}
Here, we describe the datasets used in our experiments and the experimental setup. Furthermore, we provide a quantitative and qualitative analysis of the results.

\subsection{Datasets}
We selected three sign language datasets designed for ISLR covering three different sign languages, American, Turkish, and Mexican sign language.

\subsubsection{WLASL (WLASL-100 subset)}
\begin{figure*}[ht]
 \includegraphics[width=\linewidth]{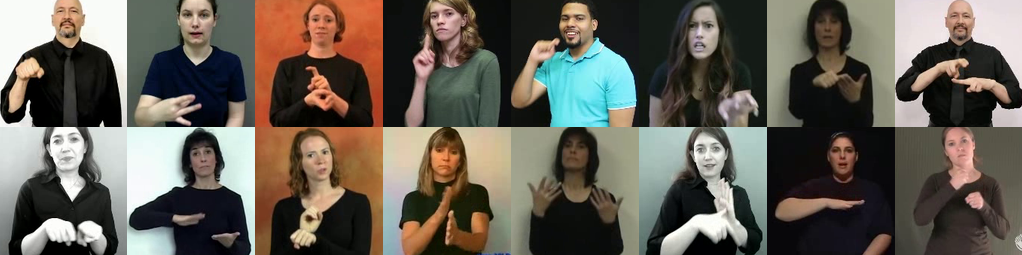}
 \centering
 \caption{Illustration of the WLASL dataset containing videos with multiple backgrounds, illumination conditions, and signers with different appearances.}
 \label{wlasl_illustration}
\end{figure*}

The Word-Level American Sign Language dataset (WLASL) \cite{li2020word} is a large-scale dataset of isolated ASL videos. It contains 2,000 unique classes distributed across 21,083 videos and 119 unique signers. In each video, the signer performs a single sign in a nearly-frontal view. The videos are collected from 20 different websites including ASLU, ASL-LEX, and YouTube, providing a very diverse amount of videos with different backgrounds and lighting conditions, as shown in Figure \ref{wlasl_illustration}. The WLASL dataset is divided into 3 subsets, WLASL-100, WLASL-300, and WLASL-2000. The WLASL-100 subset contains 100 classes distributed across 2038 videos and 119 unique signers. This subset is split into train, validation, and testing sets. 1442 videos by 91 unique signers for training, 338 videos by 69 unique signers for validation, and 258 videos by 56 unique signers for testing. The videos are decoded with 25 fps (frames per second) and resized to 256x256 pixels. In this work, we used the WLASL-100 subset to estimate the skeleton sequences.

\subsubsection{AUTSL (RGB data track)}
The Ankara University Turkish Sign Language dataset (AUTSL) \cite{sincan2020autsl} is a large-scale and diverse collection of isolated Turkish Sign Language (TSL) videos, comprising 226 signs performed by 43 distinct signers, with a total of 36,302 video samples. It contains 20 different backgrounds and includes signers who are deaf, coda, TSL instructors, TSL translators, TSL students, and trained individuals.
The dataset is split into train, validation, and testing sets. 28,142 videos for training, 4,418 videos for validation, and 3,742 videos for testing. The dataset was recorded using Microsoft Kinect v2 and contains the RGB data of the videos in addition to the depth. In this work, we used only the RGB data track to estimate the skeleton sequences.

\subsubsection{LSM Dataset (Mejía-Pérez \cite{mejia2022automatic})}
This dataset consists of 3,000 individual sign language samples, covering 30 unique signs of Mexican Sign Language (LSM). Each sign was performed 25 times by four different signers. The signs were recorded using an OAK-D camera, with 20 consecutive frames captured for each sign. At each frame, 543 keypoints from the face, body, and hands were extracted using MediaPipe. A subset of 67 keypoints is provided as follows: 20 for the face, 5 for the body, and 21 for each hand. The coordinates of the keypoints were transformed to meters using the focal length and the depth data captured by the camera. The coordinates were normalized with respect to the inner chest to compensate for the variations in the distance between the camera and the signer. We performed a slight modification to the base graph depicted in Figure \ref{fig:skeleton_graph_representation} to account for the 67 keypoints excluding the nose.

\subsection{Experimental Setup}
\begin{figure*}[ht]
 \includegraphics[width=\linewidth]{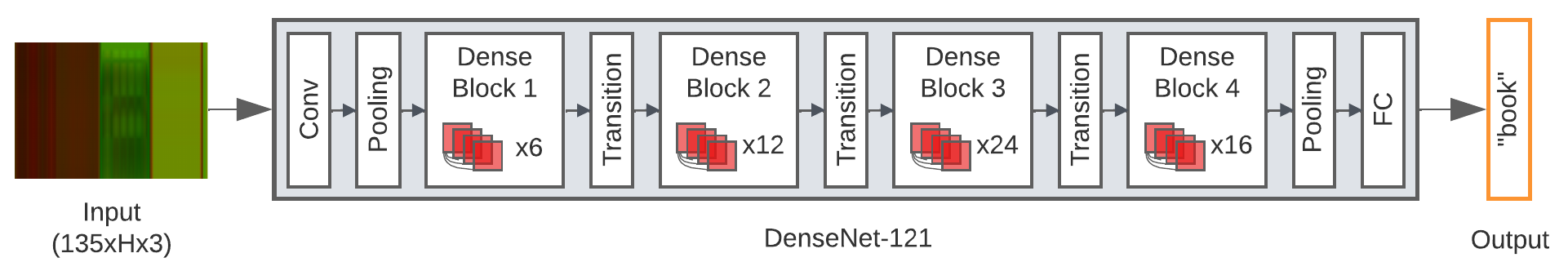}
 \centering
 \caption{The TSSI representation of a skeleton sequence with dimensions 135xHX3 (width, height, channels) is processed by a DenseNet-121 network to output a predicted label for classification. ``H" is set to the mean sequence length in the training set.}
 \label{deep_learning_model}
\end{figure*}

We tested the proposed input representation TSSI with a very well-known deep learning architecture for image classification, DenseNet-121 \cite{huang2017densely}, as shown in Figure \ref{deep_learning_model}. This network is designed to improve feature reuse and gradient flow through the use mainly of 4 dense blocks of multiple layers connected densely, each block separated by a transition layer that performs down-sampling via convolution and pooling, ending with a global pooling and a fully connected layer. We used the implementation available in the Keras library \cite{chollet2015keras}. We added dropout before the last layer to boost the generalization performance and modified the number of units of the last layer to match the number of classes of the datasets.


The experiments were carried out using an NVIDIA DGX workstation with a V100 GPU. We used a stratified 5-fold cross-validation strategy to perform hyperparameter tuning. Then, we trained the model with the training and validation set of the datasets using its best hyperparameters. We used 100 epochs for the WLASL-100 dataset and 24 epochs for the AUTSL and the LSM dataset. Finally, we evaluated the models on the test set to measure the performance. At every training, we used the cross-entropy loss and the stochastic gradient descent with Nesterov momentum and momentum = 0.98 for optimization. We used the pre-trained weights in the ImageNet dataset as initialization \cite{deng2009imagenet} except when training on the AUTSL and the LSM dataset.

For hyperparameter tuning, we followed the procedure proposed by \cite{smith2018disciplined} that uses learning rate range tests to select the learning rate range and other hyperparameters such as weight decay, dropout, and batch size for a cyclical learning rate schedule. We performed a grid search of the following hyperparameter configurations: batch size = [32, 64], dropout = [0.1, 0.3, 0.5], weight decay = [1e-5, 1e-6, 1e-7], learning rate range = (0.001 - 1.0). Table \ref{table:hyperparameters} shows the final hyperparameters used in each dataset.

\begin{table}[ht]
\centering
\begin{tabular}{@{}lcccc@{}}
\toprule
\textbf{Dataset} & \textbf{BS} & \textbf{WD} & \textbf{DO} & \textbf{LRR} \\
\midrule
WLASL-100 & 64 & 1e-5 & 0.3 & 0.001-0.0065\\
AUTSL & 64 & 1e-5 & 0.5 & 0.01-0.5 \\
LSM & 64 & 1e-5 & 0.3 & 0.01-0.1 \\
\bottomrule
\end{tabular}
\caption{Hyperparameters used for each dataset. ``BS" : batch size, ``WD" : weight decay, ``DO" : dropout, ``LRR" : learning rate range.}
\label{table:hyperparameters}
\end{table}

\subsection{Quantitative Results}
We report the categorical top-1 accuracy achieved on the testing sets of the WLASL-100, AUTSL, and the LSM dataset. The results are compared to existing skeleton-based and RGB-based methods excluding multi-modal architectures.

\begin{table}[ht]
\centering
\begin{tabular}{@{}llr@{}}
\toprule
\textbf{Method} & \textbf{Input} & \textbf{Accuracy} \\
\midrule
I3D (baseline) \cite{li2020word} & \multirow{3}{*}{RGB} & 65.89 \\
TK-3D ConvNet \cite{li2020transferring}  & & 77.55 \\
Full Transformer Network  \cite{du2022full} & & 80.72 \\
\midrule
GCN-BERT \cite{tunga2021pose} & \multirow{3}{*}{Skeleton} & 60.15 \\
Pose-TGCN \cite{li2020word} & & 55.43 \\
SPOTER  \cite{bohavcek2022sign} & & 63.18 \\
\midrule
SL-TSSI-DenseNet (ours) & \multirow{2}{*}{Skeleton} & 73.02 \\
\textbf{SL-TSSI-DenseNet (ours) + DA} & & \textbf{81.47} \\
\bottomrule
\end{tabular}
\caption{Comparison of the state-of-the-art in the WLASL-100 dataset including our model. ``Input" : input modality. ``RGB": Raw RGB videos as input. ``Skeleton": it uses any form of skeleton data as input. ``DA": "Data Augmentation".}
\label{table_model_benchmark}
\end{table}

\begin{table}[ht]
\centering
\begin{tabular}{@{}llr@{}}
\toprule
\textbf{Method} & \textbf{Input} & \textbf{Accuracy} \\
\midrule
CNN + FPM + BLSTM \\ + Attention (baseline) \cite{sincan2020autsl} & \multirow{3}{*}{RGB} & 49.22 \\
I3D + RGB-MHI \cite{sincan2022using} & & 93.53 \\
ResNet2 + 1D \cite{jiang2021skeleton} & & 95.00 \\
\textbf{SlowFast + Slow + TSM} \\ \textbf{(wenbinwuee team)} \cite{sincan2021chalearn} & & \textbf{96.55}\\
\midrule
Multi-stream SL-GCN \\ (2D Keypoints) \cite{jiang2021skeleton} & \multirow{1}{*}{Skeleton} & 96.47 \\
SSTCN \cite{jiang2021skeleton} & & 93.37 \\
\midrule
SL-TSSI-DenseNet (ours) & Skeleton & 93.13 \\
\bottomrule
\end{tabular}
\caption{Comparison of the state-of-the-art in the AUTSL dataset including our model. ``Input" : input modality. ``RGB": Raw RGB videos as input. ``Skeleton": it uses any form of skeleton data as input.}
\label{table_autsl_model_benchmark}
\end{table}

\begin{table}[ht]
\centering
\begin{tabular}{@{}llr@{}}
\toprule
\textbf{Method} & \textbf{Input} & \textbf{Accuracy} \\
\midrule
RNN (baseline) \cite{mejia2022automatic} & \multirow{3}{*}{Skeleton} & 92.44 \\
LSTM \cite{mejia2022automatic} & & 96.66\\
GRU \cite{mejia2022automatic} & & 97.11 \\
\midrule
\textbf{SL-TSSI-DenseNet (ours)} & \textbf{Skeleton} & \textbf{98.0} \\
\bottomrule
\end{tabular}
\caption{Comparison of the state-of-the-art in the LSM dataset including our model. ``Input" : input modality. ``Skeleton": it uses any form of skeleton data as input.}
\label{table_mejiaperez_model_benchmark}
\end{table}

\subsubsection{WLASL-100}
Table \ref{table_model_benchmark} shows that our model achieves better results than other skeleton-based models and competitive results against RGB-based models. Some of the most well-known state-of-the-art models for SLR to which we compare treat skeletons as graphs, and others introduce them in transformers architectures. We overcome models such as GCN-BERT \cite{tunga2021pose}, SPOTER \cite{bohavcek2022sign}, and Pose-TGCN \cite{li2020word}.
GCN-Bert processes a skeleton sequence as a graph with a graph convolutional neural network (GCN) to model the spatial relationships and BERT \cite{devlin2018bert} to learn temporal representations. Pose-TGCN introduces the temporal dimensionality in a GCN to also process an entire skeleton sequence as a graph. SPOTER receives a skeleton sequence as a vector and proposes a slight modification of the original Transformer \cite{vaswani2017attention} by feeding the decoder with the class representation of the sample to process it.

Similarly, when compared to RGB-based input representations, SL-TSSI-DenseNet overcomes models such as I3D which uses 52M parameters while our model only uses 7.2M parameters. However, our model falls short when compared to Full Transformer Network \cite{du2022full} and TK-3D ConvNet \cite{li2020transferring}, probably for the difference in complexity represented by the size of the models. For instance, Full Transformer Network works with 20M parameters, and TK-3D ConvNet with 52M parameters, approximately. After adding data augmentation, our SL-TSSI-DenseNet overcomes all RGB-based and skeleton-based models.

\subsubsection{AUTSL (RGB data track)}
The results obtained on the AUTSL dataset are compared to other skeleton-based and RGB-based methods in Table \ref{table_autsl_model_benchmark}.
Even though our model does not overcome the other skeleton-based methods it presents competitive results with 93.13\% accuracy while keeping a lower number of parameters. For instance, Multi-stream SL-GCN \cite{jiang2021skeleton} uses around 19.2M parameters and employs spatio-temporal graph convolutional modules to process four graph representations of a skeleton sequence based on the joints and the bones vectors.
On the other hand, the RGB-based model proposed by the wenbinwuee team for the ChaLearn LAP Large Scale Signer Independent Isolated Sign Language Recognition Challenge \cite{sincan2021chalearn} uses at least 33M parameters as it processes the RGB data using SlowFast \cite{feichtenhofer2019slowfast}, SlowOnly \cite{feichtenhofer2019slowfast} and TSM \cite{lin2019tsm} independently and fuses the class scores at the end to produce a final prediction.

\subsubsection{LSM dataset (Mejía-Pérez)}
The results obtained in the LSM dataset are presented in Table \ref{table_mejiaperez_model_benchmark}. The baseline methods proposed by \cite{mejia2022automatic}, RNN, LSTM, and GRU take the skeleton data as a vector input and involve the use of recurrent dropout and a dense layer at the end of the network. Our model overcomes these models with 98.0\% test accuracy. This dataset has not been benchmarked by any other model yet.

\begin{figure}
  \centering
  \begin{subfigure}{\linewidth}
    \includegraphics[width=\linewidth]{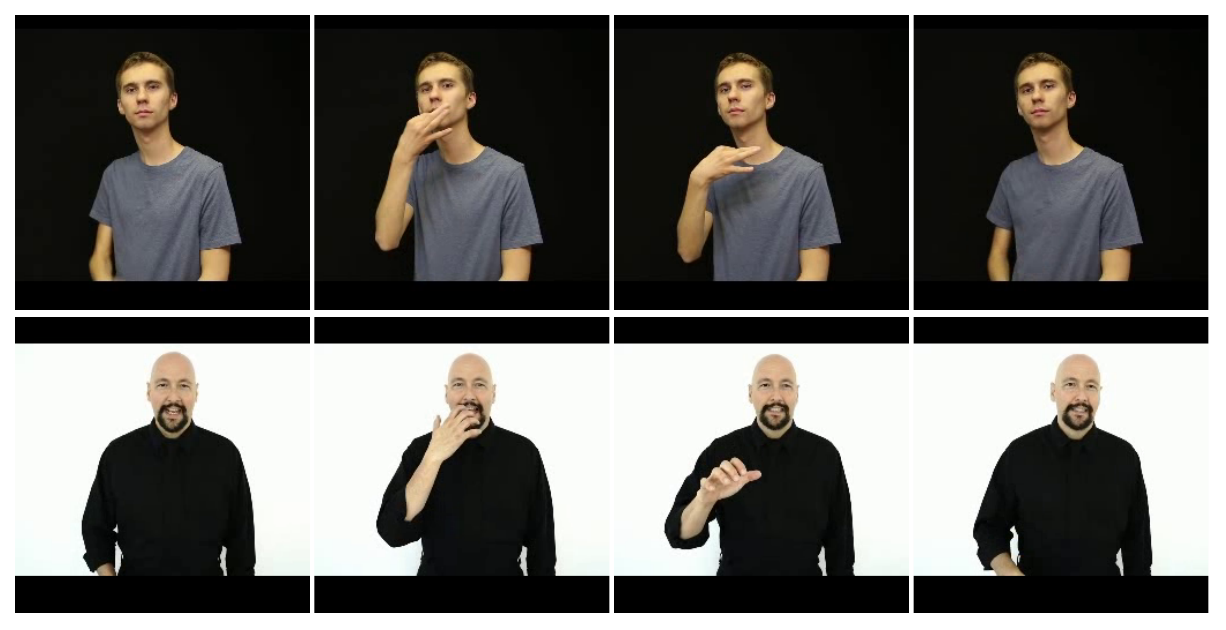}
    \caption{``thin" (upper) vs ``hot" (lower)}
    \label{fig:short-a}
  \end{subfigure}
  \vfill
  \begin{subfigure}{\linewidth}
    \includegraphics[width=\linewidth]{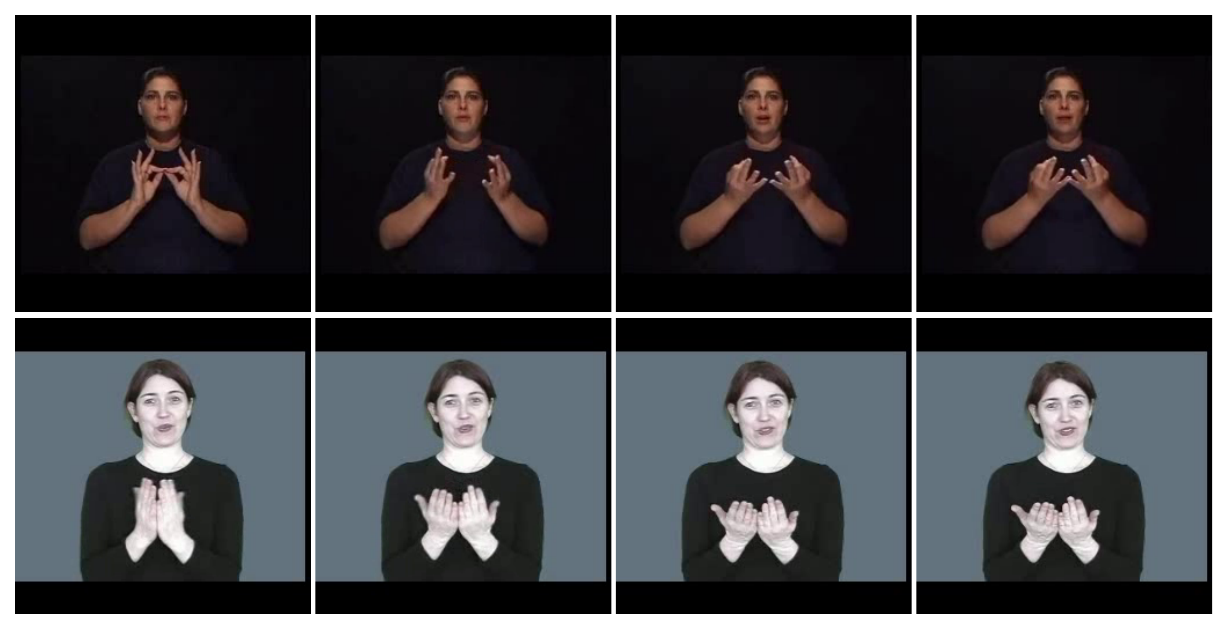}
    \caption{``family" (upper) vs ``book" (lower)}
    \label{fig:short-b}
  \end{subfigure}
  \vfill
  \begin{subfigure}{\linewidth}
    \includegraphics[width=\linewidth]{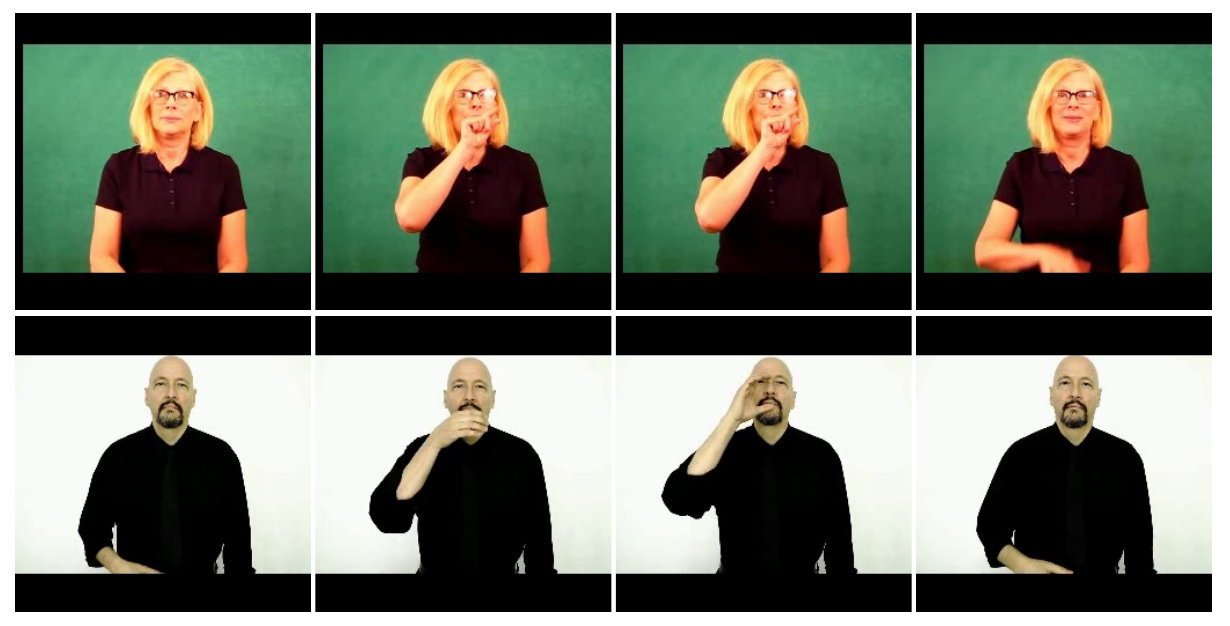}
    \caption{``bird" (upper) vs ``drink" (lower)}
    \label{fig:short-c}
  \end{subfigure}
  \caption{Similar signs predicted incorrectly.}
  \label{fig:similar_signs}
\end{figure}

\subsection{Qualitative Results}
The qualitative analysis of the results is based on the results obtained on the WLASL-100 dataset. We obtained the confusion matrix of the testing set and visualized the signs along with the sign with which they were mostly misclassified. As shown in Figure \ref{fig:similar_signs}, signs such as ``thin", ``family" and ``bird" were misclassified as ``hot", ``book" and ``drink", respectively. This might be due to that the signs are performed with similar hand positions and shapes. This leads us to enhance the capability of the model to pay attention more specifically to the hand shapes through attention modules or other mechanisms.

\subsection{Ablation study}
We performed an ablation study using the WLASL-100 dataset to evaluate the effect of pre-training and data augmentation of the model. We used 3 data augmentation techniques that transform the spatial and temporal characteristics of the skeleton motion: 1) \textit{Scale}, scales the skeleton by a random factor between 0.5 and 1.0 to mimic different body sizes, 2) \textit{Flip}, flips horizontally the skeleton with a random probability of 0.5, and 3) \textit{Speed}, resizes vertically the TSSI to a random number of frames between 48 (25th percentile) and the 74 (75th percentile) of the training set video length using bilinear resizing.

As shown in Table \ref{table:model_configurations}, a baseline model (A) trained without pre-training or data augmentation achieves 39.15\% accuracy. By adding only data augmentation, model (B) obtains an increase of around 19\% in accuracy. By adding only pre-training, model (C) obtains an increase of around 34\% in accuracy. Finally, by adding both pre-training and data augmentation, the model (D) obtains an increase of around 42\% in accuracy. The results show that the accuracy in WLASL-100 increases with pre-training despite the fact that the pre-trained weights come from the ImageNet dataset, which belongs to a different domain than sign language.

\begin{table}[ht]
\centering
\begin{tabular}{@{}c|cccc@{}}
\toprule
\textbf{Model}              &
\textbf{Pre-training}       &
\textbf{Augmentation}               &
\textbf{Accuracy} \\
\hline
A       &
\xmark       &
\xmark       &
39.15 \\
B       &
\xmark       &
\cmark       &
58.45 \\
C       &
\cmark       &
\xmark       &
\textbf{73.02} \\
D       &
\cmark       &
\cmark       &
\textbf{81.47} \\
\bottomrule
\end{tabular}
\caption{Average top-1 accuracy on 5 runs of models generated with different configurations.}
\label{table:model_configurations}
\end{table}

Table \ref{table_ablation} shows the results of an ablation study to determine the effects of the data augmentation techniques in the best model obtained with pre-training and data augmentation. The results show that the speed augmentation technique is the most important as the accuracy drops down to 65.82\% when it is removed. It also shows that the flip augmentation and the scale augmentation do not have a substantial impact when they are removed as the accuracy drops by only 1

\begin{table}[ht]
\centering
\begin{tabular}{@{}lcccc@{}}
\toprule
\textbf{DA Technique}           &
\textbf{None}               &
\textbf{Flip}               &
\textbf{Speed}              &
\textbf{Scale} \\
\midrule
\textbf{Accuracy }    &
\textbf{81.47}              &
80.78                       &
65.82                       &
81.25 \\
\bottomrule
\end{tabular}
\caption{Average top-1 accuracy on 5-fold cross-validation after removing individually the data augmentation techniques ``Flip", ``Speed" and ``Scale". The column ``None" represents the result of not removing any data augmentation technique.}
\label{table_ablation}
\end{table}

\section{Conclusions}
\label{sec:conclusions}
We showed that our proposed approach, SL-TSSI-DenseNet, which uses TSSI to convert a skeleton sequence into an image and process the image with a DenseNet-121, represents an alternative for isolated sign language representation. Our model offers superior performance than other skeleton-based models in the WLASL-100 dataset. Furthermore, using data augmentation, it can overcome RGB-based models while being less complex, in terms of size. We validated the effectiveness of our approach in two other datasets, the AUTSL dataset, and the LSM dataset. In the AUTSL dataset, we achieved a competitive performance in comparison to other skeleton-based methods. In the LSM dataset, a recent-published dataset that has not been benchmarked yet by other models, we obtained a higher performance than the baseline. Future work can explore the effects of using weights pre-trained on HAR or SLR datasets instead of the ImageNet dataset. Additionally, future work can explore the effects of adding an attention mechanism to the model and using depth data.
\section*{Acknowledgments}

The authors wish to thank the AI Hub at Tecnologico de Monterrey for their support for carrying out the experiments reported in this paper on their NVIDIA's DGX computer. We also wish to thank CONACYT for the master's scholarship for David Laines at Tecnologico de Monterrey.

{\small
\bibliographystyle{ieee_fullname}
\bibliography{egbib}
}

\end{document}